\documentclass[runningheads]{llncs}
\usepackage[T1]{fontenc}
\usepackage[pagebackref=true,breaklinks=true,letterpaper=true,colorlinks,bookmarks=false]{hyperref}
\usepackage{amssymb,amsmath,array}
\usepackage{graphicx}
\usepackage{svg}
\definecolor{yellow}{RGB}{221,170,51}
\definecolor{red}{RGB}{187,85,102}
\definecolor{blue}{RGB}{0, 68, 136}
\usepackage{subcaption}

\begin{document}
\title{Automated Semantic Rules Detection (\textsc{ASRD}) for Emergent Communication Interpretation}
\titlerunning{\textsc{ASRD} for Emergent Communication Interpretation}
%
\author{Bastien Vanderplaetse\inst{1}\inst{*}\orcidID{0000-0002-6542-7334} \and
Xavier Siebert\inst{2}\orcidID{0000-0003-0869-7968} \and
Stéphane Dupont\inst{1}\orcidID{0000-0003-3674-6747}}
\authorrunning{B. Vanderplaetse et al.}

%
%
\institute{University of Mons, MAIA Artificial Intelligence Lab, Mons, Belgium
\email{\{bastien.vanderplaetse,stephane.dupont\}@umons.ac.be}\\
\and
University of Mons, Mathematics and Operational Research, Mons, Belgium\\
\email{xavier.siebert@umons.ac.be}\\
* Corresponding author}
\maketitle              
\begin{abstract}
The field of emergent communication within multi-agent systems examines how autonomous agents can independently develop communication strategies, without explicit programming, and adapt them to varied environments. However, few studies have focused on the interpretability of emergent languages. The research exposed in this paper proposes an Automated Semantic Rules Detection (\textsc{ASRD}) algorithm, which extracts relevant patterns in messages exchanged by agents trained with two different datasets on the Lewis Game, which is often studied in the context of emergent communication. \textsc{ASRD} helps at the interpretation of the emergent communication by relating the extracted patterns to specific attributes of the input data, thereby considerably simplifying subsequent analysis.

\end{abstract}
\section{Introduction}
The study of emergent communication in multi-agent systems aims to uncover how autonomous agents can independently create and refine their own interaction protocols. This research area connects computational linguistics, artificial intelligence, and robotics, focusing on the development of complex communication strategies among autonomous agents. Depending on the task they are trained for, these agents may cooperate or be in competition with one another. To communicate, agents need a protocol communication, which refers to a set of rules, conventions, and structures that govern how information is exchanged between agents in a multi-agent system. These protocols define the format, timing, sequencing, and semantics of messages, ensuring that agents can effectively interpret and respond to the information they receive. More specifically, emergent communication protocols focus on the functional aspects of communication, such as how messages are encoded, transmitted, and decoded. These protocols govern the mechanics of interaction, ensuring that agents can exchange information efficiently and reliably. Communication protocols require a language, which refers to the symbolic system or vocabulary that agents develop to represent and convey meaning. It encompasses the creation of symbols, grammar, and shared understanding, allowing agents to express complex ideas and concepts. While emergent communication protocols deal with the "how" of communication, emergent language deals with the "what" and the shared understanding that emerges from interactions.

In traditional systems, communication protocols and languages are often explicitly designed and predefined, tailored to specific tasks or environments. However, in the context of emergent communication, either the protocol, the language, or both are not specified, but instead evolve dynamically as agents interact and learn from their environment and each other. This emergent process allows for the development of flexible and adaptive communication strategies, which can be particularly advantageous in complex or novel scenarios where predefined protocols and languages may be insufficient or impractical. However, learned protocols and languages raise challenges in interpretability. This has driven interest in using discrete symbols and syntactic structures inspired by human language, valued for its adaptability and ability to convey intricate ideas across new contexts. Emergent communication thus proposes a solution to create scalable, adaptable discrete communication languages that could hopefully mirror human linguistic structures to improve interpretability. However, the interpretation of messages exchanged by the agents in the context of emergent communication has not yet been extensively studied. This paper proposes an interpretation method based on an algorithm, Automated Semantic Rules Detection (\textsc{ASRD}), that extracts patterns occuring in the emergent language between the agents. This algorithm identifies constant positions in messages by grouping them based on attribute or hyperattribute values representing the properties of the observed data. It then extracts representative patterns by removing global invariants, linking these patterns to the data properties. Our results show that \textsc{ASRD} is able to find relevant patterns in the messages used by agents and to determine their relation to attributes defining the input data observed by the agents. 

Section~\ref{sec:relatedworks} reviews related works in the field of emergent communication, highlighting the gaps in interpretability and the need for automated analysis tools.

Section~\ref{sec:emcomm} provides a detailed explanation of emergent communication, focusing on the fundamental concepts of vocabulary, messages, and language in multi-agent systems.

Section~\ref{sec:algo} introduces the \textsc{ASRD} algorithm, describing its workflow and how it extracts semantic rules from emergent communication protocols.

Section~\ref{sec:exp} outlines the experimental setup, including the agents' architecture, image feature transformations, datasets, and evaluation metrics used in this paper.

Section~\ref{results} presents the results of the experiments, analyzing the performance of the agents and the effectiveness of \textsc{ASRD} in extracting semantic rules from emergent languages.

Section~\ref{sec:ccl} discusses the implications of the findings, highlights the limitations of the current approach, and suggests directions for future research.

\section{Related Works}\label{sec:relatedworks}
Emergent communication builds on the idea that, when agents share an environment and work towards common objectives, they can develop communication strategies without human intervention~\cite{foerster_learning_2016}. Recent advances have highlighted the flexibility of these emergent protocols, showcasing their ability to improve coordination and collaboration across diverse tasks and scenarios~\cite{masquil_intrinsically-motivated_2022,wang_emergence_2022}. Current research primarily focuses on improving the overall performance of multi-agent systems, with additional emphasis on the resilience and adaptability of emergent protocols under various conditions~\cite{bullard_quasi-equivalence_2021,wang_emergence_2022}. While emergent protocols improve agent collaboration, a notable gap in the literature lies in the interpretability of emergent languages and their alignment with human understanding, which is a critical aspect for ensuring the usability and trust of such systems in real-world contexts~\cite{zhu_survey_2024}. Some research works propose to design adaptive languages that enhance the clarity and sparsity of messages to meet the needs of diverse human-agent teams~\cite{karten_interpretable_2023}. Others works explore the development of communication strategies more aligned with human language~\cite{bouchacourt_how_2018}, specifically on its compositionality property~\cite{mordatch_emergence_2018}. However, the interpretation of language emerging from such techniques requires a lot of manual analysis. 

In this paper, we propose a method to help at this interpretation thereby minimizing manual analysis. To focus on the language emergence and its interpretation, we test our algorithm on a simple task involving two agents with a simple communication protocol: the Lewis Game, often used in studies of emergent communication~\cite{chaabouni_emergent_2022,lazaridou_emergence_2018}.

\section{Emergent Communication}\label{sec:emcomm}
Emergent communication in multi-agent systems (\textsc{MAS}) refers to the process by which autonomous agents develop their own communication protocol or language without explicit instructions. In this paper, we focus on emergent language. Therefore, this Section defines the fundamental concepts of an emergent language.

\subsection{Vocabulary and Messages}
In \textsc{MAS}, communication between agents requires a \textit{vocabulary} $\mathcal{V}$, which is a finite, non-empty set of symbols $\left\{w_1, \ldots, w_n\right\}$, where $n \in \mathbb{N}_0$ represents the number of symbols. Each element $w_i\in\mathcal{V}$ is referred to as a \textit{word}. The set of all possible finite sequences of words from $\mathcal{V}$ is denoted as $\mathcal{V}^*$.

Agents use the words from $\mathcal{V}$ to construct messages. A \textit{message} $m = \left(w_1,\ldots,w_T\right)$ is a finite sequence of words, where each $w_i \in \mathcal{V}$ and $T$ is the length of the message. The set of all possible messages is denoted as $\mathcal{M} \subseteq \mathcal{V}^*$. Within a message, each instance of a word from $\mathcal{V}$ is called a \textit{token}.

\subsection{Language in Multi-Agent Systems}
A \textit{language} $\mathcal{L}$ over a vocabulary $\mathcal{V}$ is a structured system of communication used by agents to convey information. Formally, a language is defined as a set of messages selected from $\mathcal{V}^*$, along with a set of rules that govern how these messages are constructed and interpreted. Specifically, a language can be represented as $\mathcal{L} = \left(\mathcal{V}, \mathcal{R}_\mathcal{L}\right)$, where $\mathcal{R}_\mathcal{L}$ is the set of syntactic and semantic rules that define how symbols can be combined and understood.

The set of syntactic rules, called the \textit{syntax}, is denoted as $\mathcal{R}_\texttt{syntax}$ and refers to the rules that determine the structure and order of words in a message. These rules specify how words from $\mathcal{V}$ can be combined to form valid messages.

The set of semantics rules, called the \textit{semantics}, refers to the meaning assigned to words and messages. It is represented as a function $f_M: \mathcal{M} \rightarrow \mathcal{I}$, where $\mathcal{I}$ is the set of possible meanings. This function maps each message $m\in\mathcal{M}$ to its corresponding interpretation or meaning.

\subsection{Emergent communication}
In emergent communication, only the vocabulary $\mathcal{V}$ is predefined. The agents must collaboratively construct the set of rules $\mathcal{R}_\mathcal{L}$—including both syntax $\mathcal{R}_\texttt{syntax}$ and semantics $f_M$—during the training process. Through interaction and coordination, agents develop a shared understanding of how to combine words into meaningful messages and how to interpret these messages in context.

Once the training phase is complete, we can analyze the emergent language to understand its syntax and semantics. This is an important step to interprete how agents communicate. This paper proposes \textsc{ASRD}, an algorithm aiming to extract semantic rules from the emergent language developed between two agents on a specific use case: the Lewis Game.

\section{Automated Semantic Rules Detection}\label{sec:algo}
\textsc{ASRD} is an algorithm designed to detect semantic rules in the messages exchanged by agents, with the goal of linking these rules to specific attributes and hyperattributes derived from the input data. In this work, the input data consists of images observed by the agents. By identifying these patterns, ASRD provides a crucial first step toward interpreting the emergent communication strategies developed by the agents.

\subsection{Task: Lewis Game}
The Lewis Game~\cite{lewis_convention_1969} serves as the experimental framework for many studies in emergent communication~\cite{chaabouni_emergent_2022,lazaridou_emergence_2018}. It is a referential communication task involving two collaborating agents: the \textit{Speaker} and the \textit{Listener}. The game proceeds in three phases: 
\begin{enumerate}
    \item \textbf{Message Generation:} The \textit{Speaker} receives a target image $x$ and generates a message $m$.
    \item \textbf{Message Interpretation:} The \textit{Listener} interprets the message $m$ and selects an image $\hat{x}$ from a candidate set $\mathcal{C}$, which includes the target image $x$.
    \item \textbf{Reward:} Both agents receive a reward based on whether the selected image $\hat{x}$ matches the target image $x$.
\end{enumerate}
 The goal of the game is for the agents to develop an effective language that allows the \textit{Listener} to accurately identify the target image based on the \textit{Speaker}'s message.

\subsection{Attributes and Hyperattributes}
In the context of the Lewis Game, attributes refer to the fundamental properties that define an image. These are explicit, directly observable features that uniquely characterize an image. For example, in an image dataset, attributes might include shape, color, or spatial relationships between objects.

Hyperattributes, on the other hand, are higher-order properties derived from the attributes. They represent abstract or boolean characteristics that generalize across multiple images. For instance, a hyperattribute might indicate whether an image contains a specific combination of shapes or whether certain spatial relationships are present. Hyperattributes are particularly useful for analyzing higher-level patterns in agent communication, as they abstract away specific attribute combinations and highlight broader relationships between message structure and image properties.

The specific attributes and hyperattributes used in our experimental setup are described in details in Section~\ref{sec:attrdef}.

\subsection{Algorithm Workflow}
\textsc{ASRD}\footnote{\href{https://github.com/bastienvanderplaetse/ASRD}{https://github.com/bastienvanderplaetse/ASRD}} operates through a series of well-defined steps aimed at identifying and analyzing the relationships between message tokens and the attributes or hyperattributes of the input data. The workflow consists of four key steps:
\begin{enumerate}
    \item \textbf{Data Grouping:} The dataset is partitioned into groups based on the value of a selected attribute or hyperattribute. For each group, all corresponding messages are retrieved. This step ensures that the analysis is conducted independently for each property value.
    \item \textbf{Identification of Constant Positions:} Within each group, the algorithm scans all retrieved messages to identify constant positions—token positions where the value remains the same across all messages in the group. These positions are considered indicative of a relationship between the token values and the attribute or hyperattribute.
    \item \textbf{Combination Analysis for Multiple Positions:} In cases where multiple token positions consistently vary together, the algorithm treats these positions as a combination. The corresponding token sequences are recorded as potential markers of the attribute or hyperattribute value.
    \item \textbf{Global Constant Removal:} Token positions that remain constant across all groups, regardless of the attribute or hyperattribute value, are excluded from the analysis. This step eliminates global invariants—token values that do not contribute to distinguishing between different property values.
\end{enumerate}

The output of ASRD is a list of patterns, where each pattern is defined by a unique combination of symbols at identified token positions. Each pattern is associated with a set of attribute and hyperattribute values that are common to all images described by messages containing this pattern. These patterns correspond to semantic rules and provide insights into how agents encode and communicate information about the input data.

\section{Experimental Setup}\label{sec:exp}
In this study, we use agents' architecture from existing litterature~\cite{chaabouni_emergent_2022} and trained based on the work of~\cite{vanderplaetse_influence_2024}, who used different transformations on the ouput of the image encoders for these architectures. Specifically, we focus on the ResNet-18 architecture~\cite{he_deep_2015}, while using the three image feature transformation methods: $\mathcal{F}_1$ (average pooling), $\mathcal{F}_2$ (flattening), and $\mathcal{F}_3$ (flattening with custom initialization of the weights of specific layers). Below, we provide a detailed explanation of the agents and the methods used.

\subsection{Agents architecture}
The agents used in our experiments are based on the Speaker and Listener architectures described in recent works~\cite{chaabouni_emergent_2022,vanderplaetse_influence_2024}. These agents were trained on the Lewis Game, a referential communication task involving two agents: the Speaker, which generates messages based on a target image, and the Listener, which interprets the message to select the correct image from a set of candidates. The training process employed a population-based approach~\cite{chaabouni_emergent_2022} to avoid overfitting and extreme co-adaptation, which is common in single-pair agent setups~\cite{lanctot_unified_2017}. In this work, we use a population of 10 \textit{Speakers} and 10 \textit{Listeners}.

\begin{figure}
\includegraphics[width=\textwidth]{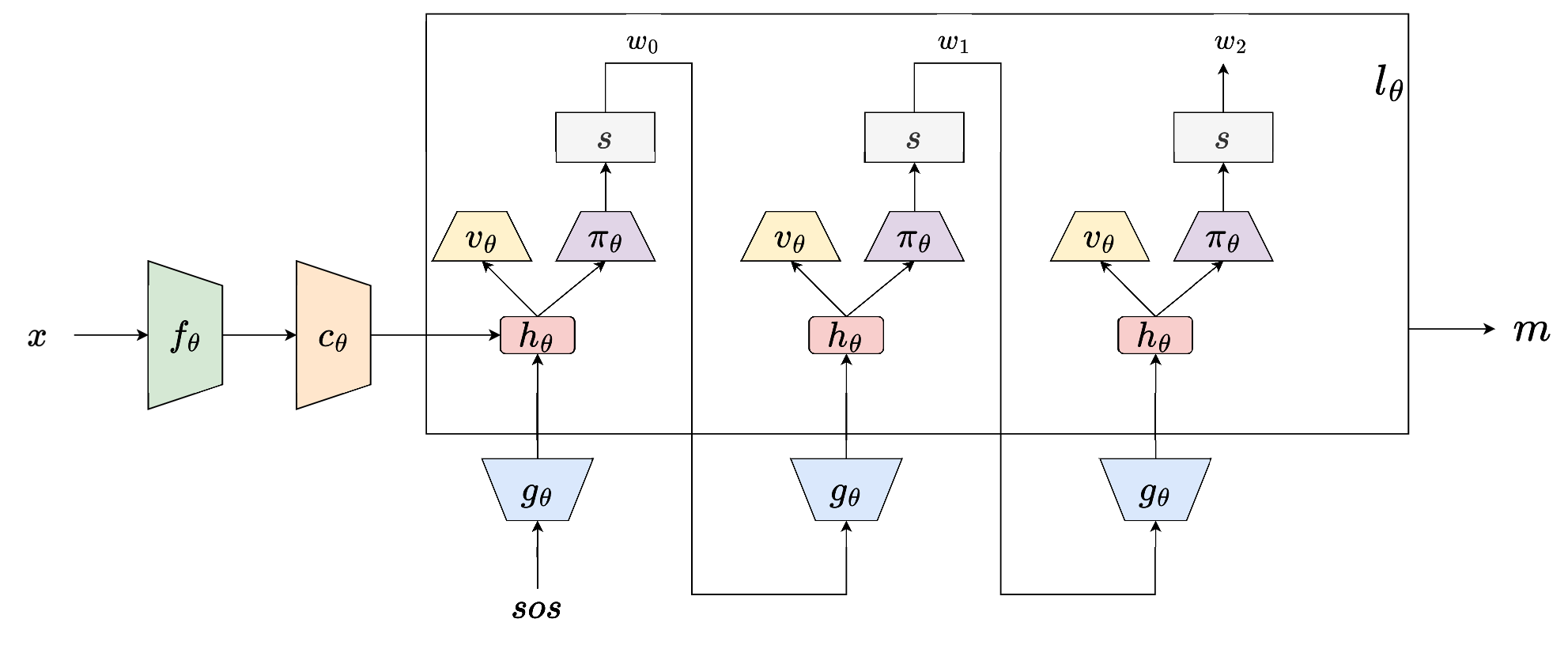}
\caption{\textit{Speaker}'s architecture~\cite{chaabouni_emergent_2022,vanderplaetse_influence_2024}.} \label{fig:speaker}
\end{figure}

The \textit{Speaker} is a neural network designed to generate a message $m$ corresponding to a given image $x$. As illustrated in Figure~\ref{fig:speaker}, \textit{Speaker} comprises multiple modules. First, an image encoder $f_\theta$ processes the image $x$ and maps it to an embedding representation $f_\theta(x)$. This embedding is then projected by a state adapter $c_\theta$ into the initial hidden state of a long short-term memory (\textsc{LSTM}) network~\cite{hochreiter_long_1997}, such that $z_{-1,\theta} = c_\theta(f_\theta(x))$.  

Subsequently, the LSTM module $l_\theta$ iteratively updates its hidden state based on the input word embedding $e_{t,\theta} = g_\theta(w_{t-1})$, where $g_\theta$ denotes an embedding layer. The generation process is initialized using a special \textit{start-of-sequence} (\texttt{sos}) token, yielding the initial embedding $e_{0,\theta} = g_\theta(\texttt{sos})$.  

At each time step $t$, the \textsc{LSTM} state $z_{t,\theta} = l_\theta(z_{t-1,\theta}, e_{t,\theta})$ is used to compute outputs through two distinct linear layers: (i) a value head $v_\theta$, which estimates the expected reward, and (ii) a policy head $\pi_\theta$, which determines the probability distribution over the vocabulary for the next word $w_t$. During training, the next word $w_t$ is sampled from this distribution using a sampling function $s$, such that $w_t = s(\pi_\theta(.|z_{t,\theta}))$. At inference time, the word selection is performed greedily. The generated word $w_t$ is subsequently fed back into $g_\theta$ to compute the next word embedding, and this process continues iteratively until a predefined maximum message length $T$ is reached.

\begin{figure}
\includegraphics[width=\textwidth]{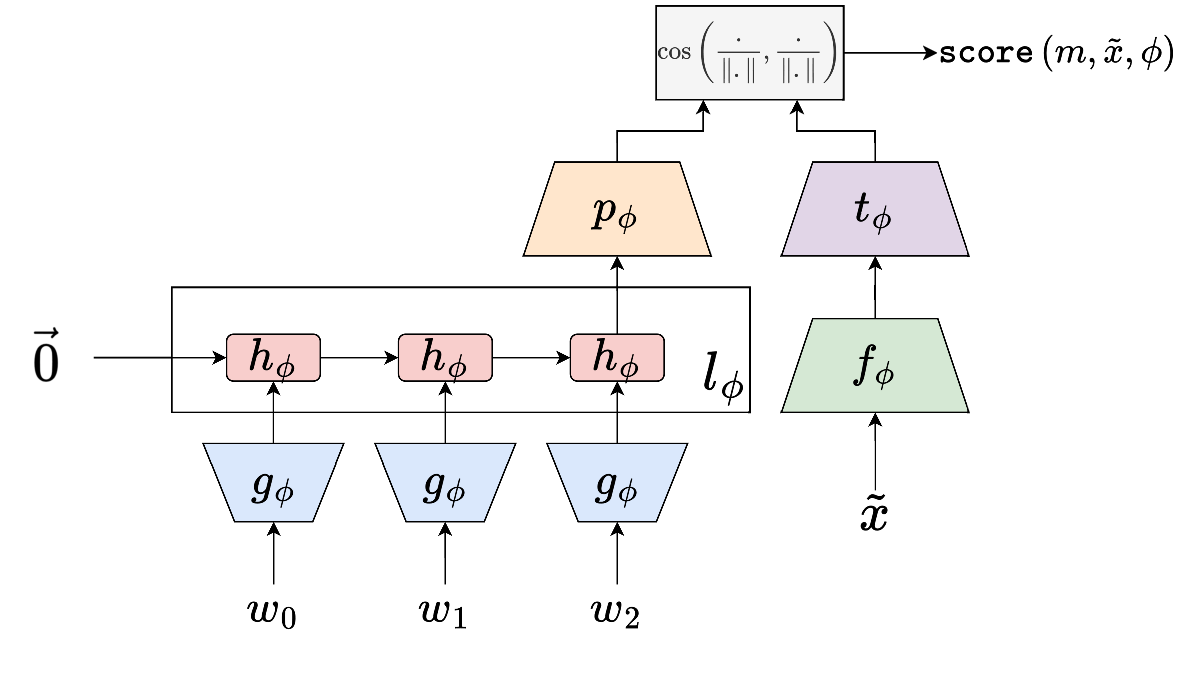}
\caption{\textit{Listener}'s architecture~\cite{chaabouni_emergent_2022,vanderplaetse_influence_2024}.} \label{fig:listener}
\end{figure}

The \textit{Listener} is a neural network tasked with interpreting the message $m$ generated by \textit{Speaker} and identifying the target image $x$ from a set of candidate images $\mathcal{C}$, which includes $x$. As illustrated in Figure~\ref{fig:listener}, \textit{Listener} computes a probability distribution over the candidate images, estimating the likelihood that a given image $\Tilde{x} \in \mathcal{C}$ corresponds to the target $x$.  

To process the message, \textit{Listener} sequentially encodes the word embeddings $e_{t,\phi} = g_\phi(w_t)$ using a \textsc{LSTM} network $l_\phi$, which updates its state according to $z_{t,\phi} = l_\phi(z_{t-1,\phi}, e_{t,\phi})$. The LSTM is initialized with a null vector $z_{-1,\phi}$, and its final hidden state $z_{T-1,\phi}$ is passed through a linear projection layer $p_\phi$, referred to as the core-state projection network, to produce the message representation $p_{m,\phi} = p_\phi(z_{T-1,\phi})$.  

In parallel, each candidate image $\Tilde{x} \in \mathcal{C}$ is processed through an image encoder $f_\phi$ to obtain an embedding $f_\phi(\Tilde{x})$. This embedding is subsequently projected via a linear layer $t_\phi$, referred to as the target projection network, yielding the final image representation $t_{\Tilde{x}} = t_\phi(f_\phi(\Tilde{x}))$.  

To assess the similarity between the message and each candidate image, a score function is defined as  
\begin{equation*}
    \texttt{score}(m,\Tilde{x}, \phi) = \sum_i\left(\frac{p_{m,\phi}}{\left\Vert p_{m,\phi}\right\Vert_2} - \frac{t_{\Tilde{x}}}{\left\Vert t_{\Tilde{x}}\right\Vert_2}\right)_i^2.
\end{equation*}

A softmax function is then applied to normalize the scores across all candidate images, producing a probability distribution $\pi_\phi(.|m,\mathcal{C})$. Finally, \textit{Listener} selects the image $\hat{x}$ that maximizes this probability:  
\begin{equation*}
\hat{x} \in \arg\max_{\Tilde{x} \in \mathcal{C}} \pi_\phi(\Tilde{x} | m, \mathcal{C}).
\end{equation*}

\subsection{Image Feature Transformations}
In this work, we use the three methods applied by~\cite{vanderplaetse_influence_2024} to the image features extracted from the ResNet-18 encoders $f_\theta$ and $f_\phi$:
\begin{itemize}
    \item \textbf{$\mathcal{F}_1$:} This method applies average pooling to the extracted image features, reducing their spatial dimensions. Average pooling is a common technique for downsampling feature maps, but it can lose fine-grained spatial details.
    \item \textbf{$\mathcal{F}_2$:} This method flattens the feature maps into a single vector by concatenating all spatial dimensions. Unlike average pooling, it preserves the complete spatial structure of the features, maintaining a more detailed representation of the image.
    \item \textbf{$\mathcal{F}_3$:} This method flattens the image features into a single vector and applies a custom initialization to the weights of the subsequent layer ($c_\theta$ for the \textit{Speaker} and $t_\phi$ for the \textit{Listener}). Specifically, each feature value $f_i$ is connected to the $i$th neuron of the next layer with a weight of $1$, while all other connections are set to $0$. This approach retains the spatial information of the features and ensures that each feature value is directly mapped to a specific neuron in the next layer.
\end{itemize}

\subsection{Datasets}
For our experiments, we use two image datasets.

\begin{figure}[h!]
\centering
\includegraphics[width=0.3\columnwidth]{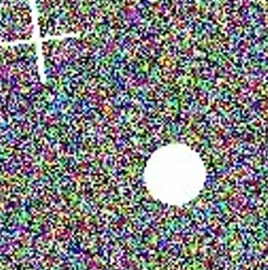}
\caption{Image generated for a combination ($\bullet$, $\times$, $\nwarrow$) from \textsc{MOPRD}.}\label{fig:moprd}
\end{figure}

\paragraph{1. Multi-Object Positional Relationships Dataset (\textsc{MOPRD}).} \textsc{MOPRD}~\cite{feng_learning_2023} is composed of images with 100 unique combinations of object attributes (\textit{shape}, \textit{shape}, \textit{relationship}). The shapes can be \textit{square} ($\square$), \textit{circle} ($\circ$), \textit{filled square} ($\blacksquare$), \textit{filled circle} ($\bullet$) or \textit{cross} ($\times$), while the relationship can be \textit{right} ($\rightarrow$), \textit{top-right} ($\nearrow$), \textit{top} ($\uparrow$) or \textit{top-left} ($\nwarrow$). Figure~\ref{fig:moprd} presents an example of an image from \textsc{MOPRD}. 80 out of 100 combinations are used for training and 20 for testing. This split ensure the evaluation of agents on combinations they have not encountered during training, thereby assessing their ability to develop generalization through compositionality in the emergent language structures. To test our algorithm, we generate 100 images for each combination and we process each \textit{Speaker} on these images to retrieve a set of messages. If a \textit{Speaker} generates different messages for the same combination (\textit{shape}, \textit{shape}, \textit{relationship}), we only keep the messages for which the number of instances represents at least 15\% of all the messages for this combination. This selection allows to consider only the most used messages for each combination. The rejected messages are considered as outliers since they do not represent how the \textit{Speaker} represents the information retrieved from the images.

\paragraph{2. Visual Genome Human-Animal-Circular (\textsc{VGHAC}).} \textsc{VGHAC} is based on a specific sampling over the Visual Genome (\textsc{VG}) dataset. \textsc{VG} is designed to enhance image understanding by providing detailed annotations for a large number of images~\cite{krishna_visual_2017}. It contains over 108,000 images, each richly labeled with region descriptions, objects, attributes, and relationships between objects, resulting in millions of distinct annotations. 

To create \textsc{VGHAC}, we decided to select a restraint number of objects from \textsc{VG}. First, for each object $o$, we removed all the attributes having less than 100 instances when associated with $o$. This ensures that the attributes of all selected objects will have a high enough number of instances in \textsc{VGHAC}. Next, we wanted to select objects respecting several conditions:
\begin{itemize}
    \item A high number of instances, which ensures that the dataset contains enough samples for this object.
    \item A large variety of attributes, which ensures there is a compositionality to learn in the image structure.
    \item The selected objects may be grouped into different categories.
\end{itemize}

Based on these criteria, we selected 6 types of objects, grouped into three categories:
\begin{enumerate}
    \item Humans: \textit{man}, \textit{woman}, \textit{person};
    \item Animals: \textit{bear}, \textit{cat}, \textit{giraffe};
    \item Circular objects: \textit{pizza}, \textit{plate}, \textit{wheel}.
\end{enumerate}

\begin{figure}[h!]
\centering
\centering
    \begin{subfigure}[t]{0.3\textwidth}
        \centering
        \includegraphics[height=1in]{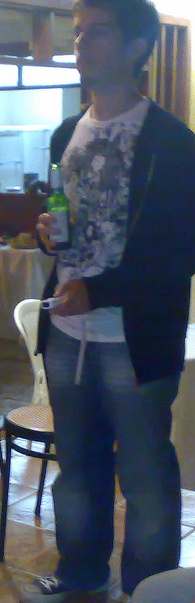}
        \caption{Entity \textit{man}.}
    \end{subfigure}%
    ~ 
    \begin{subfigure}[t]{0.3\textwidth}
        \centering
        \includegraphics[height=1in]{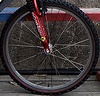}
        \caption{Entity \textit{wheel}.}
    \end{subfigure}
    ~ 
    \begin{subfigure}[t]{0.3\textwidth}
        \centering
        \includegraphics[height=1in]{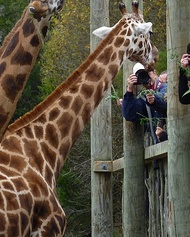}
        \caption{Entity \textit{giraffe}.}
    \end{subfigure}
\caption{Sample images from \textsc{VGHAC}.}\label{fig:vghac}
\end{figure}

For each selected object, we crop the image following the object's bounding box. Figure~\ref{fig:vghac} presents some sample images of these cropped images. This result in a total of 96 different attributes. The resulting images are split into a training set of 48,226 images and a test set of 16,661 images. 

\subsection{Attributes and Hyperattributes Definition}\label{sec:attrdef}
\paragraph{\textsc{MOPRD}} An image from \textsc{MOPRD} being defined by a tuple (\textit{shape1}, \textit{shape2}, \textit{relationship}), the attributes are trivial: \textit{shape1}, \textit{shape2} and \textit{relationship}. A straightforward definition of the hyperattributes would be to consider if the shapes in the image are filled or not. Consequently, we define the hyperattributes \textit{fill1} (resp. \textit{fill2}) as boolean being \textit{True} if \textit{shape1} (resp. \textit{shape2}) is $\blacksquare$ or $\bullet$. \textit{fill1} and \textit{fill2} can be combined to form the hyperattributes \textit{all\_fill} and \textit{all\_empty}. Finally, we define the boolean hyperattribute \textit{aligned}, which is \textit{True} if \textit{relationship} is \textit{right} or \textit{top}. This hyperattribute is used to differentiate positioning that are horizontal or vertical from the diagonale ones.


\paragraph{\textsc{VGHAC}} A trivial attribute for an image from \textsc{VGHAC} is the type of object represented on the image. From this attribute, we can define an hyperattribute \textit{group\_entity} which is \textit{human}, \textit{animal} or \textit{circular}. With the 96 entities attributes extracted from Visual Genome, we define a list of 18 \textsc{ASRD} attributes (e.g., \textit{position}, \textit{age} or \textit{color}), whose values are the attribute from Visual Genome.

\subsection{Evaluation metrics}
\paragraph{Topographic Similarity (TopSim)} is a widely used metric for evaluating the compositionality of emergent languages in multi-agent communication systems~\cite{oord_representation_2019}. TopSim measures the alignment between the structure of the input space (e.g. images) and the message space by computing the Spearman correlation between distances in the two spaces. A high TopSim value indicates that the emergent language exhibits compositional properties, where similar inputs produce similar messages, and dissimilar inputs produce dissimilar messages.

TopSim has been extensively applied in studies of emergent communication to assess the generalization capabilities of agents and the structure of the languages they develop~\cite{lazaridou_emergence_2018,li_ease--teaching_2019}. It is particularly useful for evaluating how well agents encode and communicate the underlying features of the input data, such as object shapes, colors, and spatial relationships. 

In this paper, we follow the approach of prior works~\cite{chaabouni_emergent_2022,feng_learning_2023} and use the edit distance in the message space. This distance is based on the Levenshtein distance, which measures the distance by counting the number of token changes required to transform one message into another. For the image space, two approaches are usually applied: (i) the cosine distance between the image features obtained by the image encoder of the agents, and (ii) the edit distance, which consists of counting the number of attributes with different values between the two images. 

Since \textsc{ASRD} is based on the attributes defining the images, we use the edit distance in the image space. By using this measure, a high value of \textit{TopSim} means that the emergent language is based on the attributes of the image. Therefore, \textsc{ASRD} should detect meaningful patterns in the emergent language.

\paragraph{Accuracy} is used for evaluating the performance of agents in the Lewis Game. It measures how effectively the Listener can correctly identify the target image based on the message generated by the Speaker. In this paper, we evaluate each possible (\textit{Speaker},\textit{Listener}) pair. The accuracy for a \textit{Speaker} $S_i$ corresponds to the average of the accuracy the \textit{Listeners} get when paired with $S_i$. High accuracy indicates that the \textit{Speaker} is able to generate informative messages and that the \textit{Listener} can reliably interpret these messages to identify the correct image.

\section{Results}\label{results}
The results presented in Table~\ref{tab:resume_result} provide an overview of the performance of the agents in the Lewis Game, as well as the effectiveness of \textsc{ASRD} in extracting semantic rules from the emergent communication protocols. The analysis is conducted across two datasets, \textsc{MOPRD} and \textsc{VGHAC}, and three different image feature transformation methods: $\mathcal{F}_1$ (average pooling), $\mathcal{F}_2$ (flattening), and $\mathcal{F}_3$ (flattening with custom initialization).

\begin{table}[]
\centering
\caption{Accuracy, topographic similarity and number of semantic rules extracted by \textsc{ASRD} for each \textit{Speaker} with the different image feature extraction methods on both \textsc{MOPRD} and \textsc{VGHAC}.}
\label{tab:resume_result}
\begin{tabular}{l|c|c|c|c}
\textbf{Dataset}                & \textbf{$\mathcal{F}_i$} & \textbf{Accuracy} &\textbf{TopSim}          & \textbf{\#rules for each \textit{Speaker}} \\ \hline
{MOPRD} & $\mathcal{F}_1$ & $23.51\pm 0.14$ &$12.59\pm 0.07$ & 6 (for all)       \\
                       & $\mathcal{F}_2$ & $35.99\pm 0.12$ & $21.88\pm 0.13$ & 8 (for all)       \\
                       & $\mathcal{F}_3$ & $27.44 \pm 0.02$ & $24.39\pm 0.11$ & 10 (for all)      \\ \hline
{VGHAC} & $\mathcal{F}_1$ &  $95.38\pm 0.12$   & $0.12\pm 0.08$                & 21; 21; 21; 24; 22; 23; 19; 21; 21; 20        \\
                       & $\mathcal{F}_2$ &  $94.96\pm 0.27$   & $0.18$                & 21; 17; 19; 25; 21; 19; 18; 22; 21; 21         \\
                       & $\mathcal{F}_3$ & $94.06\pm 0.15$   & $0.00\pm 0.05$               &  15; 14; 15; 14; 17; 15; 12; 15; 15; 15       
\end{tabular}
\end{table}

\subsection{Performance Metrics}
\paragraph{Accuracy:} The accuracy metric measures the ability of the Listener to correctly identify the target image based on the message generated by the Speaker. For the \textsc{MOPRD} dataset, the accuracy varies across the different feature transformation methods. As can be seen in Table~\ref{tab:resume_result}, the highest accuracy is achieved with $\mathcal{F}_2$, followed by $\mathcal{F}_3$ and $\mathcal{F}_1$. This suggests that preserving the spatial structure of the image features through flattening ($\mathcal{F}_2$) leads to more informative messages, enabling the Listener to make more accurate identifications. In contrast, average pooling ($\mathcal{F}_1$) appears to lose some of the fine-grained details necessary for accurate communication.

For the \textsc{VGHAC} dataset, the accuracy is consistently high across all transformation methods, with $\mathcal{F}_2$ achieving the highest accuracy, closely followed by $\mathcal{F}_2$ and $\mathcal{F}_3$. This indicates that the agents are highly effective in communicating and interpreting messages for this dataset, regardless of the feature transformation method used. The high accuracy in \textsc{VGHAC} may be attributed to the richer and more diverse images available in this dataset, which allows an easier distinction between images.

\paragraph{TopSim:} TopSim measures the alignment between the structure of the input space (images) and the message space, indicating the compositionality of the emergent language. For the \textsc{MOPRD} dataset, $\mathcal{F}_3$ achieves the highest TopSim value, followed by $\mathcal{F}_2$ and $\mathcal{F}_1$. This suggests that the custom initialization in $\mathcal{F}_3$ enhances the compositionality of the emergent language, making it more structured and interpretable. In contrast, the low TopSim value for $\mathcal{F}_1$ indicates that average pooling may hinder the development of a compositional language.

For the \textsc{VGHAC} dataset, the TopSim values are significantly lower, revealing that the messages are not correlated with the attributes defining the different images of the dataset.

\subsection{ASRD analysis}
The number of semantic rules extracted by \textsc{ASRD} provides insights into the complexity of the emergent language. For the \textsc{MOPRD} dataset, the number of rules is consistent across all Speakers for each image feature transformation method. This indicates that $\mathcal{F}_3$ leads to a more complex and nuanced communication protocol, which is consistent with the higher TopSim values observed for this method.

In Table~\ref{tab:18f1s0}, we present the identified patterns for the \textit{Speakers} trained using ResNet-18 and the transformation method $\mathcal{F}_1$\footnote{The results for others experiments are available here: \textit{anonymous}}. First, we observe that all the messages have 7 fixed values, giving the general pattern 13-12-\textbf{\textcolor{yellow}{XX}}-10-10-10-10-10-\textbf{\textcolor{red}{YY}}-\textbf{\textcolor{blue}{ZZ}}. Only 6 words have thus been used in the emerging language, which we shorten the messages in \textbf{\textcolor{yellow}{XX}}-\textbf{\textcolor{red}{YY}}-\textbf{\textcolor{blue}{ZZ}} for concision. With these patterns extracted by \textsc{ASRD}, we can determine the message that will be generated for a given image. For example, if both shapes are $\circ$ and if the hyperattributes \textit{fill1}, \textit{fill2} and \textit{all\_fill} are \textit{False}, the message will be \textbf{\textcolor{yellow}{12}}-\textbf{\textcolor{red}{YY}}-\textbf{\textcolor{blue}{ZZ}}. However, if the shapes are $\square$ or $\times$, we know that the message is \textbf{\textcolor{yellow}{12}}-\textbf{\textcolor{red}{10}}-\textbf{\textcolor{blue}{ZZ}}. With this rule, \textsc{ASRD} also detects that both shapes are empty (\textit{all\_empty} is \textit{True}). About the filled shapes, \textsc{ASRD} detects the patterns \textbf{\textcolor{yellow}{XX}}-\textbf{\textcolor{red}{YY}}-\textbf{\textcolor{blue}{10}} when both shapes are $\blacksquare$ and \textbf{\textcolor{yellow}{XX}}-\textbf{\textcolor{red}{10}}-\textbf{\textcolor{blue}{10}} when they are $\bullet$. To indicate that \textit{all\_fill} is \textit{True}, which implies that the shapes are $\blacksquare$ or $\bullet$, while being different, the message is \textbf{\textcolor{yellow}{10}}-\textbf{\textcolor{red}{10}}-\textbf{\textcolor{blue}{10}}. This shows that, even if the \textit{Speaker} is not able to tell which shape is $\blacksquare$ or $\bullet$, it still conveys the information that both shapes are filled, which proves the utility for hyperattributes. With \textsc{ASRD}'s output, we also observe that no message conveys information about the relationship between the shapes, since the last row of Table~\ref{tab:18f1s0} shows that, for any value of the relationship between shapes, we do not have any information of the token used to represent the \textit{relationship} attribute.

\begin{table}[h!]
\caption{Recognized patterns for experiments with ResNet-18 and method $\mathcal{F}_1$ for \textsc{MOPRD}.}
\label{tab:18f1s0}
\centering
\resizebox{\textwidth}{!}{
\begin{tabular}{ccccccccccc}
\multicolumn{3}{l|}{\textbf{Pos. in message}}                                   & \multicolumn{3}{l|}{\textbf{Attributes}}                                                                                                                         & \multicolumn{5}{l}{\textbf{Hyperattributes}}                                                                                                \\ \hline
\multicolumn{1}{c|}{\textcolor{yellow}{\textbf{2}}}  & \multicolumn{1}{c|}{\textcolor{red}{\textbf{8}}}  & \multicolumn{1}{c|}{\textcolor{blue}{\textbf{9}}}  & \multicolumn{1}{c|}{\textbf{shape1}}             & \multicolumn{1}{c|}{\textbf{shape2}}             & \multicolumn{1}{c|}{\textbf{relationship}}                                            & \multicolumn{1}{c|}{\textbf{fill1}} & \multicolumn{1}{c|}{\textbf{fill2}} & \multicolumn{1}{c|}{\textbf{all fill}} & \multicolumn{1}{c|}{\textbf{all empty}} & \textbf{aligned} \\ \hline
\multicolumn{1}{c|}{12} & \multicolumn{1}{c|}{}   & \multicolumn{1}{c|}{}   & \multicolumn{1}{c|}{$\circ$}            & \multicolumn{1}{c|}{$\circ$}            & \multicolumn{1}{c|}{}                                               & \multicolumn{1}{c|}{F}     & \multicolumn{1}{c|}{F}     & \multicolumn{1}{c|}{F}        & \multicolumn{1}{c|}{}          &         \\ \hline
\multicolumn{1}{c|}{12} & \multicolumn{1}{c|}{10} & \multicolumn{1}{c|}{}   & \multicolumn{1}{c|}{$\square$ $\times$} & \multicolumn{1}{c|}{$\square$ $\times$} & \multicolumn{1}{c|}{}                                               & \multicolumn{1}{c|}{}      & \multicolumn{1}{c|}{}      & \multicolumn{1}{c|}{}         & \multicolumn{1}{c|}{T}         &         \\ \hline
\multicolumn{1}{c|}{}   & \multicolumn{1}{c|}{}   & \multicolumn{1}{c|}{10} & \multicolumn{1}{c|}{$\blacksquare$}     & \multicolumn{1}{c|}{$\blacksquare$}     & \multicolumn{1}{c|}{}                                               & \multicolumn{1}{c|}{T}     & \multicolumn{1}{c|}{T}     & \multicolumn{1}{c|}{}         & \multicolumn{1}{c|}{F}         &         \\ \hline
\multicolumn{1}{c|}{}   & \multicolumn{1}{c|}{10} & \multicolumn{1}{c|}{10} & \multicolumn{1}{c|}{$\bullet$}          & \multicolumn{1}{c|}{$\bullet$}          & \multicolumn{1}{c|}{}                                               & \multicolumn{1}{c|}{}      & \multicolumn{1}{c|}{}      & \multicolumn{1}{c|}{}         & \multicolumn{1}{c|}{}          &         \\ \hline
\multicolumn{1}{c|}{10} & \multicolumn{1}{c|}{10} & \multicolumn{1}{c|}{10} & \multicolumn{1}{c|}{}                   & \multicolumn{1}{c|}{}                   & \multicolumn{1}{c|}{}                                               & \multicolumn{1}{c|}{}      & \multicolumn{1}{c|}{}      & \multicolumn{1}{c|}{T}        & \multicolumn{1}{c|}{}          &         \\ \hline
\multicolumn{1}{c|}{}   & \multicolumn{1}{c|}{}   & \multicolumn{1}{c|}{}   & \multicolumn{1}{c|}{}                   & \multicolumn{1}{c|}{}                   & \multicolumn{1}{c|}{$\rightarrow$ $\nearrow$ $\uparrow$ $\nwarrow$} & \multicolumn{1}{c|}{}      & \multicolumn{1}{c|}{}      & \multicolumn{1}{c|}{}         & \multicolumn{1}{c|}{}          & F, T    \\ \hline
\multicolumn{11}{l}{Pattern message: 13-12-{\color{yellow}{XX}}-10-10-10-10-10-{\color{red}{YY}}-{\color{blue}{ZZ}}}                                                                                     
\end{tabular}}
\end{table}

For the \textsc{VGHAC} dataset, the number of rules varies more significantly across Speakers. However, for this dataset, a higher number of rules does not correlate with a better compositionality. This can be illustrated for instance, for one of the \textit{Speakers} trained with the method $\mathcal{F}_3$. In this case, \textsc{ASRD} detected that two positions are fixed in the messages and that each rule is a combination of the others to define one single value of one attribute of the image. For example, in the eight variable positions, the message \textbf{4-18-18-18-4-4-18-4} corresponds to the attribute \textit{activity} having the value \textit{skateboarding}. This behaviour shows that the language of this \textit{Speaker} tends to be holistic, in the sense that each message relates to one single concept, which corresponds to the low compositionality score.

It is important to point out that if hyperattributes are redundant with others ones, \textsc{ASRD} does not always precise their value. For example, if \textit{all\_fill} is \textit{False} (resp. \textit{True}), then \textit{all\_empty} is always \textit{True} (resp. \textit{False}), which is not visible in Table~\ref{tab:18f1s0}.

\section{Discussion and Conclusion}\label{sec:ccl}
In this paper, we introduced the Automated Semantic Rules Detection (\textsc{ASRD}) algorithm, designed to interpret emergent communication in multi-agent systems by identifying patterns in the messages exchanged by agents. Our primary contribution lies in the development of a method that significantly reduces the need for manual analysis by automatically linking message patterns to specific attributes and hyperattributes of the input data. This approach provides a crucial step towards understanding the emergent languages developed by two agents, particularly in the context of the Lewis Game.

Our experimental results highlight the effectiveness of \textsc{ASRD} in extracting meaningful patterns from emergent communication protocols. For the \textsc{MOPRD} dataset, \textsc{ASRD} identified consistent patterns across different image feature transformation methods, with the highest compositionality observed in the flattening methods. This suggests that preserving spatial information and custom initialization can lead to more structured and interpretable emergent languages. In contrast, the \textsc{VGHAC} dataset showed lower compositionality, indicating that the emergent language in this context is more holistic, with each message corresponding to a single concept rather than a compositional structure.

The accuracy metrics further support the effectiveness of the agents' communication, particularly in the \textsc{VGHAC} dataset, where high accuracy was achieved across all transformation methods. This indicates that the agents were able to develop effective communication protocols, even if the emergent language lacked strong compositional properties.

\textsc{ASRD} successfully identifies constant positions in messages and links them to specific attributes and hyperattributes of the input data. This allows for a more systematic and efficient analysis of emergent communication protocols. By extracting patterns and relating them to data properties, \textsc{ASRD} enhances the interpretability of emergent languages, making it easier to understand how agents encode and communicate information. 

The \textsc{ASRD} algorithm can be improved regarding several aspects. First, it does not always distinguish between redundant hyperattributes, such as when one hyperattribute's value is entirely determined by another. This can lead to incomplete interpretations of the emergent language. Second, in cases where the emergent language is holistic (e.g., in the \textsc{VGHAC} dataset), \textsc{ASRD} may struggle to extract meaningful patterns, as each message corresponds to a single concept rather than a combination of attributes. Finally, the effectiveness of \textsc{ASRD} is highly dependent on the quality and structure of the input data. The effectiveness of the algorithm on datasets with less structured or more complex attributes needs to be studied.

Future researches will focus on several mains tracks. The first one consists of finding an easier protocol to define attributes and hyperattributes. In the case of the \textsc{MOPRD}, this step is trivial, but with datasets with more complex images (e.g. \textsc{VGHAC}) or on others tasks than the Lewis Game, the definition of attributes and hyperattributes may be more sophisticated. The second one consists of exploring the algorithm itself to understand why the redundancy of hyperattributes prevents \textsc{ASRD} to determine the values of these hyperattributes. Finally, \textsc{ASRD} may be extended to multi-agent systems involving more than two agents or with data other than images.
\bibliographystyle{splncs04}
\bibliography{references}
%




\end{document}